\documentclass[11pt]{article}
\usepackage{pifont}
\newcommand{\cmark}{\ding{51}}
\newcommand{\xmark}{\ding{55}}
\usepackage{pifont}
\renewcommand{\cmark}{\ding{51}}
\usepackage{xcolor}
\usepackage{pifont}
\renewcommand{\cmark}{\textcolor{green!60!black}{\ding{51}}} 
\renewcommand{\xmark}{\textcolor{red!70!black}{\ding{55}}}   
\usepackage[table]{xcolor}  
\usepackage{tabularx}
\usepackage{array}
\newcolumntype{C}{>{\centering\arraybackslash}X}
\usepackage{booktabs}
\usepackage{caption} 

\usepackage{acl}

\usepackage{times}
\usepackage{latexsym}

\usepackage{amsmath}
\usepackage[T1]{fontenc}

\usepackage[utf8]{inputenc}

\usepackage{microtype}

\usepackage{inconsolata}

\usepackage{graphicx}
\usepackage{booktabs}

\title{GCA Framework: A GCC Countries–Grounded Dataset and Agentic Pipeline for Climate Decision Support}

\author{
  \textbf{Muhammad Umer Sheikh\textsuperscript{1}},
  \textbf{Khawar Shehzad\textsuperscript{2}},
  \textbf{Salman Khan\textsuperscript{1,3}},
  \textbf{Fahad Shahbaz Khan\textsuperscript{1,4}},  \\
  \textbf{Muhammad Haris Khan\textsuperscript{1}} \\
  \textsuperscript{1}Mohamed Bin Zayed University of Artificial Intelligence (MBZUAI), UAE \\
  \textsuperscript{2}University of Missouri, USA \\
  \textsuperscript{3}Australian National University, Australia \\
  \textsuperscript{4}Linköping University, Sweden \\
  \texttt{\{muhammad.sheikh\}@mbzuai.ac.ae}
}
\usepackage{natbib}
\begin{document}
\maketitle
\begin{abstract}
Climate decision-making in the GCC states increasingly demands systems that can translate heterogeneous scientific and policy evidence into actionable guidance, yet general-purpose large language models (LLMs) remain weak both in region-specific climate knowledge and grounded interaction with geospatial and forecasting tools. We present the \textbf{GCA framework}, which unifies (i) \textbf{GCA-DS}, a curated multimodal dataset grounded in the GCC states, and (ii) \textbf{Gulf Climate Agent (GCA)}, a tool-augmented agent for climate analysis. GCA-DS comprises $\sim$200k question--answer pairs spanning governmental policies and adaptation plans, NGO and international frameworks, academic literature, and event-driven reporting on heatwaves, dust storms, and floods, complemented with remote-sensing inputs that couple imagery with textual evidence. Building on this foundation, the GCA agent orchestrates a modular tool pipeline grounded in real-time and historical signals and geospatial processing that produces derived indices and interpretable visualizations. Finally, we benchmark open and proprietary LLMs on climate tasks in the GCC states and show that domain fine-tuning and tool integration substantially improve reliability over general-purpose baselines.
\end{abstract}

\begin{figure}[t]
  \centering
  \includegraphics[width=\columnwidth]{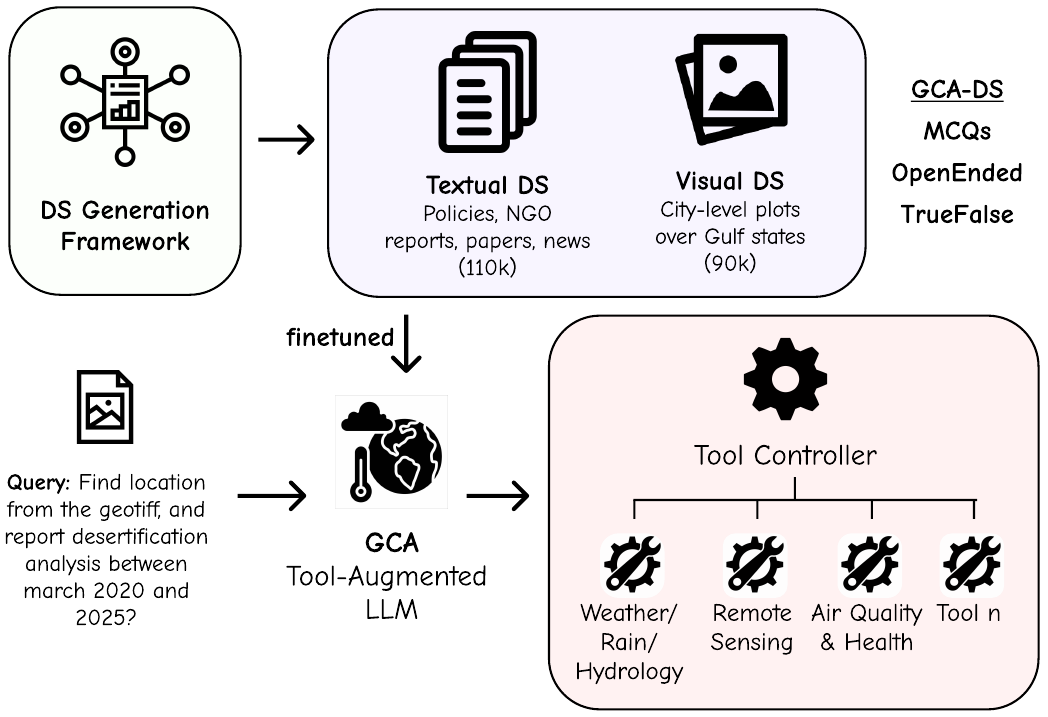}
  \caption{Overview of Gulf Climate Agent (GCA) Framework. We curate a Gulf-focused multimodal QA dataset, GCA-DS and fine-tune a tool-augmented LLM that routes user queries to specialized climate tools to produce grounded, interpretable outputs}
  \label{fig:gca-teaser}
\end{figure}

\section{Introduction}

Climate change is among the most consequential challenges for societies worldwide, yet its impacts and policy responses are profoundly shaped by regional context \citep{hewitson2014regional}. The Gulf region, encompassing the \textbf{United Arab Emirates, Saudi Arabia, Qatar, Kuwait, Oman, and Bahrain}, faces a unique constellation of climate hazards: extreme heat, dust storms, flash floods and rapid coastal erosion. These challenges demand decision support systems that can distill complex geophysical data and policy documents into actionable insights. While recent advances in large language models (LLMs) and vision--language models (VLMs) have made it possible to \emph{access and summarize} climate information at scale, general-purpose models often fall short on fine-grained tasks such as numerical reasoning over climate visualizations (trend/anomaly plots), tool-mediated retrieval of spatiotemporal variables (AQI, rainfall, discharge), and jurisdiction-specific policy interpretation. High-stakes decisions in infrastructure and policy require numerically precise, source-grounded answers that adhere to domain conventions \citep{bulian2024assessing,mastrandrea2011ipcc}, yet many models struggle with figure understanding and visual reasoning \citep{masry-etal-2022-chartqa,methani2020plotqa}. Existing climate QA benchmarks either rely on small expert-curated corpora or automatically generated datasets that suffer from noise and weak validation, and they seldom incorporate multimodal evidence or region-specific context \citep{manivannan2025climaqa}.

Within climate NLP, recent benchmarks and datasets have begun to systematize evaluation and introduce multimodal supervision. For example, \citet{bulian2024assessing} studies the adequacy of LLM responses to climate questions under a fine-grained evaluation framework; \citet{manivannan2025climaqa} propose an automated benchmark creation pipeline (ClimaGen) and expert-in-the-loop evaluation; CPIQA introduces a figure-grounded climate QA benchmark from scientific articles, targeting retrieval-augmented setups \citep{mutalik-etal-2025-cpiqa}. Despite these advances, to our knowledge there is \emph{no} publicly available resource that (i) is explicitly Gulf-grounded across policy, hazards, and geospatial evidence at large scale, and (ii) is paired with an agentic tool-augmented pipeline designed for Gulf climate objectives.

We introduce \textbf{Gulf Climate Agent (GCA)}, a framework that bridges general-purpose LLM reasoning with specialized climate tools and datasets tailored to the Gulf (see Figure \ref{fig:gca-teaser}). \textbf{Semi-automated data-generation framework.} We construct a semi-automated Gulf-specific dataset comprising approximately \textbf{200k} question--answer pairs. The dataset is sourced from government policies and adaptation strategies, NGO reports and international frameworks, academic papers on climate and sustainability, news articles describing recent heatwaves, dust storms and floods, and geospatial/remote-sensing resources (e.g., Sentinel-2 imagery and Google Earth Engine). Our data generation pipeline combines automated extraction and synthesis with human-in-the-loop verification to enable scale while maintaining reliability, grounding responses in authoritative Gulf sources rather than generic internet text.

\noindent\textbf{Agentic pipeline with climate-specific tools.} We develop an agentic pipeline that orchestrates LLM reasoning with climate-specific tool suites. Our modular architecture links the LLM to specialized modules, including heat forecasting, flood-risk prediction, carbon-footprint estimation, and coastal-erosion analysis, augmented with geospatial processing to produce interpretable maps and trends (heat maps, flood indices, air-quality trajectories, shoreline-change profiles). General-purpose retrieval and retrieval-augmented generation (RAG) components complement these modules, enabling multi-step reasoning that integrates evidence from documents and structured data.

\noindent\textbf{Benchmarking and fine-tuning.} We fine-tune and benchmark both open and proprietary LLMs (e.g., GPT-family models, Claude, Qwen) on Gulf-centric tasks, including climate question answering, policy summarisation, tool-invocation accuracy, and geospatial reasoning. Our benchmark suite includes manually annotated 91 questions and evaluation metrics for factuality, numerical precision, and tool-use reliability. Fine-tuning on the curated dataset yields domain-specialized models that improve over general-purpose baselines, highlighting the benefits of regional adaptation and tool integration.


\section{Related Work}
\label{sec:related-work}

\noindent\textbf{Climate QA and Multimodal Climate Benchmarks.}
Climate QA has evolved from machine reading over curated documents to evaluating foundation models and retrieval-augmented generation (RAG) on technical sources. Early systems such as Climate Bot \citep{rony2022climatebot} demonstrated document-grounded climate QA via CCMRC. More recent benchmarks broaden scope and difficulty: ClimaQA \citep{manivannan2025climaqa} builds expert-in-the-loop evaluation sets from graduate-level textbooks, CPIQA \citep{mutalik-etal-2025-cpiqa} introduces figure-grounded climate QA from scientific articles for RAG settings, and MMClima \citep{mmclima2026} expands multimodal coverage with expert-validated QA over figures and text. Complementarily, ClimateIQA \citep{chen2024climateiqa} targets meteorology-style visual reasoning over heatmaps. Despite these advances, existing resources are largely \emph{global} and rarely couple regional governance documents with city-level spatiotemporal evidence, which is central to Gulf decision contexts.

\begin{table*}[t]
\centering
\small
\setlength{\tabcolsep}{3pt}
\renewcommand{\arraystretch}{1.10}
\begin{tabular}{lccccccc}
\toprule
\textbf{Dataset} & \textbf{Size} & \textbf{Automated} & \textbf{Validated} & \textbf{Multimodal} & \textbf{Region} & \textbf{Remote Sensing} & \textbf{Topic Covered} \\
\midrule
Climate Crisis QA (2024) & 19,241  & \cmark & \xmark  & \xmark & General       & \xmark & 5 \\
Pirá 2.0 (2024)          & 2,250   & \xmark & \cmark  & \xmark & General       & \xmark & 4 \\
Climate-FEVER (2020)     & 1,535   & \xmark & \cmark  & \xmark & General       & \xmark & 3 \\
CPIQA (2025)             & 54,612  & \cmark & Partial & \cmark & General       & \xmark & 5 \\
ELLE (2025)              & 1,130   & \xmark & \cmark  & \xmark & General       & \xmark & 6 \\
ClimaQA-Gold (2025)      & 566     & \cmark & \cmark  & \xmark & General       & \xmark & 5 \\
ClimaQA-Silver (2025)    & 3,000   & \cmark & \xmark  & \xmark & General       & \xmark & 4 \\
MMClima (2025)           & 104,902 & \cmark & \cmark  & \cmark & General       & \xmark & 5 \\
\midrule
\textbf{GCA-DS (ours)}   & \textbf{201,410} & \cmark & Partial & \cmark & Gulf-specific & \cmark & \textbf{12} \\
\bottomrule
\end{tabular}
\vspace{2pt}
\caption{Comparison of climate- and environment-focused QA datasets. ``Partial'' denotes limited human validation (e.g., expert-guided prompting or subset review). Topic Covered reports the number of represented topic types.}
\label{tab:dataset-comparison}
\end{table*}

\noindent\textbf{Evaluating Climate Knowledge and Scientific Reasoning.}
Several studies assess LLM climate competence and scientific reliability, highlighting gaps between fluent generation and faithful, source-grounded responses \citep{bulian2024assessing,zhu2023climatechange,kufali2025climateeval}. In parallel, general scientific QA benchmarks (e.g., ScienceQA, SciQAG, SciQA) provide methodology for multimodal supervision and evaluation design, but remain domain-general and do not test climate-specific tool use or geospatial reasoning \citep{lu2022scienceqa,wan2024sciqag,auer2023sciqa}. Related work on chart and plot QA further shows persistent failure modes in numerical extraction and reasoning over scientific visualizations \citep{masry-etal-2022-chartqa,methani2020plotqa}, motivating tool-mediated computation for climate settings.

\noindent\textbf{Agentic Tool Use for Climate Workflows.}
Agentic systems increasingly operationalize tool-augmented reasoning for complex tasks. ClimateAgent \citep{kim2025climateagent} proposes a climate workflow agent and benchmark, while general tool-learning benchmarks (e.g., AgentBench, ToolLLM/ToolBench, ToolPlanner, StableToolBench) study planning, tool invocation, and evaluation stability across domains \citep{liu2024agentbench,qin2023toollm,wu2024toolplanner,guo2024stabletoolbench}. However, these works do not provide a Gulf-grounded multimodal dataset paired with a compact climate tool suite and regression-style evaluation tailored to Gulf hazards and governance.

\noindent\textbf{Summary and Positioning.}
Existing climate QA benchmarks and evaluations are primarily global and rarely integrate regional policy grounding with city-level multimodal spatiotemporal evidence (Table~\ref{tab:dataset-comparison}). To our knowledge, no prior work releases a unified Gulf-focused stack that couples (i) a large multimodal dataset with remote-sensing evidence and (ii) an agentic tool pipeline together with a dedicated tool-use regression benchmark. GCA addresses this gap by aligning Gulf-specific supervision (GCA-DS) with tool-augmented inference for decision-oriented climate queries.

\section{Curation of GCA-DS}
\label{sec:dataset}

\begin{figure}[t]
  \centering
  \includegraphics[width=\columnwidth]{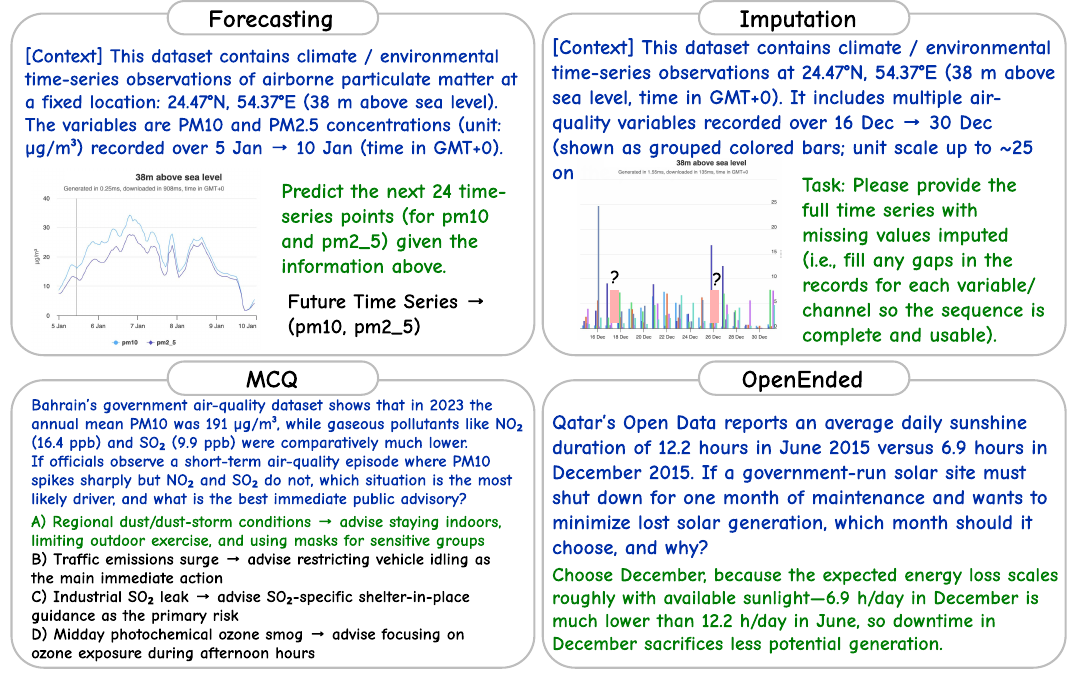}
  \caption{Example samples from the gca-ds dataset spanning text-grounded QA and visual-temporal QA over Gulf cities.}
  \label{fig:gca-dataset}
\end{figure}

\begin{figure}[t]
  \centering
  \includegraphics[width=\columnwidth]{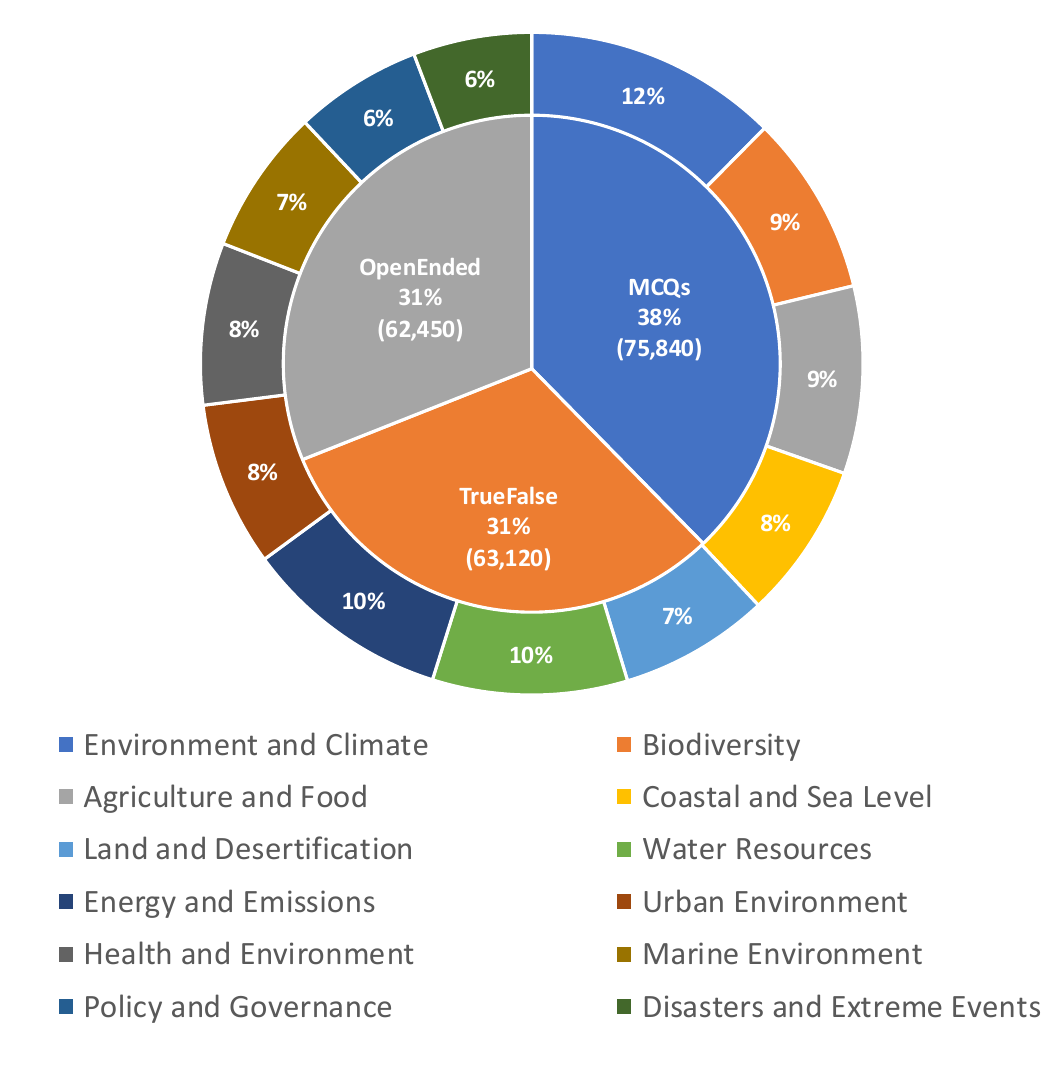}
  \caption{Distribution of GCA-DS across question types and categories.}
  \label{fig:gca-distribution}
\end{figure}

We curate a Gulf-focused multimodal climate QA dataset (GCA-DS) to support regional grounding and tool-augmented reasoning for decision-relevant queries with samples given in Figure \ref{fig:gca-dataset}. The corpus contains \textbf{200k} question--answer pairs spanning \textbf{MCQ}, \textbf{open-ended}, and \textbf{true/false} formats, with \textbf{110k} text-only items and \textbf{90k} visual-temporal items as show in Figure \ref{fig:gca-distribution}.

\paragraph{Textual data (110k).}
The textual split aggregates Gulf-relevant sources including government climate policies and adaptation strategies, NGO and international organization reports (by Gulf state), academic literature on regional climate and resilience, and news/emergency documentation covering recent extremes (heatwaves, dust storms, floods). Items are written to be evidence-grounded in the underlying documents.

\paragraph{Visual-temporal data (90k).}
The visual split is city-level across \textbf{Bahrain, Kuwait, Oman, Qatar, Saudi Arabia, and the UAE}. For each city, we construct time-indexed visual artifacts (e.g., trends and anomaly plots) from standardized meteorological and land-surface variables (ERA5/ERA5-Land) \citep{era5_cds,era5_ecmwf,era5land_cds}, atmospheric composition and air-quality variables (CAMS) \citep{cams_global_composition_ecmwf,cams_ghg_ads,cams_eu_aq_reanalysis,cams_uvindex,cams_pollen}, and hydrology signals for flooding (GloFAS) \citep{glofas_v4_reanalysis,glofas_v4_docs}. We additionally include high-resolution CMIP6 HighResMIP simulations to support scenario-style questions beyond reanalysis \citep{haarsma2016highresmip}.

\subsection{Textual Data Generation}
\label{subsec:textual-generation}

To construct the textual component of the gca-ds dataset, we employ a semi-automated pipeline that couples retrieval-grounded web acquisition with structured parsing and controlled QA synthesis. The design goal is to (i) maximize topical coverage across Gulf countries/cities and climate subdomains, while (ii) maintaining traceability to authoritative sources and reducing duplication and prompt-induced drift.

\paragraph{Retrieval-grounded keyword expansion.}
We begin by generating a set of \emph{Gulf-targeted} retrieval keys that span our four textual source classes (government climate policies, NGO reports, academic papers, and event-driven news). Given an initial seed set of topic descriptors (e.g., \emph{heatwave preparedness}, \emph{dust storm health advisory}), we use an LLM to propose candidate keywords and query templates conditioned on \emph{(country, city)} constraints. To prevent redundant crawling and near-duplicate keyword variants, we maintain a persistent vector index of previously used keywords. Each candidate keyword $k_i$ is embedded into $\mathbf{e}_i$ and filtered by its maximum cosine similarity to the existing keyword set $\mathcal{K}$:

\begin{equation}
\begin{aligned}
\mathrm{sim}(k_i,\mathcal{K})
&= \max_{k_j\in\mathcal{K}}
\frac{\mathbf{e}_i^\top \mathbf{e}_j}{\lVert \mathbf{e}_i\rVert_2 \,\lVert \mathbf{e}_j\rVert_2}, \\
k_i \text{ kept iff } \;
&\mathrm{sim}(k_i,\mathcal{K}) < \tau .
\end{aligned}
\end{equation}
This stage yields a deduplicated pool of Gulf-conditioned queries that are subsequently used for web retrieval.

\paragraph{Autonomous web retrieval agent.}
For each keyword/query, a browsing agent issues a search and iteratively refines follow-up queries when the retrieved content is off-domain or non-authoritative. This design follows the general paradigm of \emph{retrieval-augmented} systems that externalize knowledge access rather than relying solely on parametric memory \citep{lewis2020rag}. In practice, retrieval is implemented using a combination of sparse search and dense retrieval \citep{karpukhin2020dpr}, and the agent maintains a lightweight interaction trace (query $\rightarrow$ click $\rightarrow$ extract) in a ReAct-style loop that interleaves reasoning with environment actions \citep{yao2023react}. The output of this stage is a set of URLs (HTML pages and PDFs) together with minimal provenance metadata (query, timestamp, source domain).

\paragraph{Heterogeneous document parsing.}
The retrieved resources include heterogeneous formats (HTML, scanned PDFs, reports with tables/figures). We therefore normalize all sources into a unified textual representation. The parser extracts main content, removes boilerplate/navigation, preserves section headers when available, and records document metadata (title, publishing organization, date, URL) to enable later citation and auditing.

\paragraph{Semantic segmentation and chunking.}
We segment each document into overlapping chunks to preserve local coherence while controlling context length. For $D=[w_1,\dots,w_T]$, we form windows $C_m=[w_{s_m},\dots,w_{s_m+L-1}]$ with \textbf{$L=512$ tokens} and \textbf{stride $S=384$} (128-token overlap), aligning boundaries to section/paragraph breaks when available.

\paragraph{Fact induction.}
From each chunk, we induce \emph{atomic factual statements} that are designed to be verifiable and minimally compositional (one claim per statement when possible). This step reduces long-form narratives into a fact bank that supports diverse QA templates while improving grounding: each induced statement retains a pointer to its originating chunk and source metadata.

\paragraph{Instructional QA synthesis.}
Finally, we synthesize question--answer pairs in three formats, MCQ, open-ended, and true/false, by conditioning an LLM on (i) one or more atomic statements, (ii) a target question type, and (iii) a rubric that enforces answerability from the provided evidence. This follows the broader idea of self-generated instructional data creation and filtering to scale instruction tuning \citep{wang2023selfinstruct}. For MCQs, distractors are generated to be locally plausible but globally inconsistent with the evidence; for open-ended items, we enforce concise, evidence-anchored responses; and for true/false items, we include both entailed and contradicted variants to test factual robustness.

\subsection{Visual Temporal Data Generation}
\label{subsec:visual-generation}

The visual component is designed to capture \emph{temporal} and \emph{spatiotemporal} patterns that are central to Gulf climate reasoning (e.g., trend detection, anomaly identification, seasonal comparisons). We construct this split by ingesting multi-source climate and atmospheric products and rendering standardized charts that can be directly queried.

\paragraph{Multi-source climate data ingestion.}
We ingest (i) high-resolution climate simulations from CMIP6 HighResMIP models \citep{haarsma2016highresmip}, including \texttt{CMCC-CM2-VHR4}, \texttt{FGOALS-f3-H}, \texttt{HiRAM-SIT-HR}, \texttt{MRI-AGCM3-2-S}, \texttt{EC-Earth3P-HR}, \texttt{MPI-ESM1-2-XR}, and \texttt{NICAM16-8S}; (ii) atmospheric composition and aerosol forecasts from CAMS global atmospheric composition forecasts \citep{cams_global_composition_ecmwf}; (iii) air-quality reanalyses from CAMS European Air Quality Reanalysis \citep{cams_eu_aq_reanalysis}; and (iv) hydrological signals for flooding from GloFAS v4 reanalysis and forecasts \citep{glofas_v4_reanalysis,glofas_v4_forecast}. These sources support visual QA spanning meteorology (e.g., temperature, wind speed, precipitation), air quality (PM$_{10}$, PM$_{2.5}$, NO$_2$, SO$_2$, O$_3$, CO, aerosols, dust), greenhouse gases (CO$_2$, CH$_4$), health-relevant indices (UV index, pollen proxies), and hydrology (river discharge).

\paragraph{Geocoding and city inventory.}
We define a Gulf city inventory for Bahrain, Kuwait, Oman, Qatar, Saudi Arabia, and the United Arab Emirates, and map each city $c$ to its latitude--longitude pair $(\phi_c,\lambda_c)$. This enables consistent extraction from gridded products across sources.

\paragraph{Spatial index retrieval.}
Given a gridded product with coordinates $\{(\phi_i,\lambda_j)\}$, we retrieve the grid cell most relevant to a city by nearest-neighbour matching:
\begin{equation}
(i^\ast,j^\ast) = \arg\min_{i,j} \; d\big((\phi_c,\lambda_c),(\phi_i,\lambda_j)\big),
\end{equation}
where $d(\cdot,\cdot)$ is a spherical distance (or an equirectangular approximation when appropriate). This step yields a consistent time series per variable and city, and it generalizes across heterogeneous grids (0.05$^\circ$--0.1$^\circ$, model-dependent).

\paragraph{Unit and schema normalization.}
Because sources differ in naming conventions, units, time encoding, and missingness, we map each retrieved series into a canonical schema. This includes unit conversion, timestamp normalization to a unified calendar, and harmonized variable naming. Each processed record is written to a universal CSV format that serves as both (i) the canonical data representation and (ii) metadata for plot generation and QA synthesis.

\paragraph{Temporal plot builder.}
For each city--variable pair, we segment the time series into non-overlapping 3-month windows over the last 10 years. Let $\Delta = 90$ days. For window index $t$, we define:
\begin{equation}
W_t = [t\Delta, (t+1)\Delta),
\end{equation}
and drop windows where coverage is below a minimum completeness threshold (e.g., $<\rho$ observed timesteps). We then render standardized charts for each window; the corresponding CSV provides plot metadata (location, variable, units, time span, summary statistics), enabling controlled and reproducible QA generation.

\paragraph{Visual QA synthesis.}
Using the chart and its metadata, we synthesize QA items in MCQ, open-ended, and true/false formats across four primary categories: (i) anomaly detection (e.g., identifying abnormal spikes/drops), (ii) forecasting (trend continuation or seasonal projection from context), (iii) imputation (fill missing segments based on surrounding values), and (iv) reasoning QA (comparisons, aggregations, and multi-variable interpretation grounded in metadata). The resulting visual split contributes approximately \textbf{90k} QA pairs, complementing the textual split and enabling multimodal evaluation of climate reasoning in the Gulf.

\begin{figure*}[t]
  \centering
  \includegraphics[width=\textwidth]{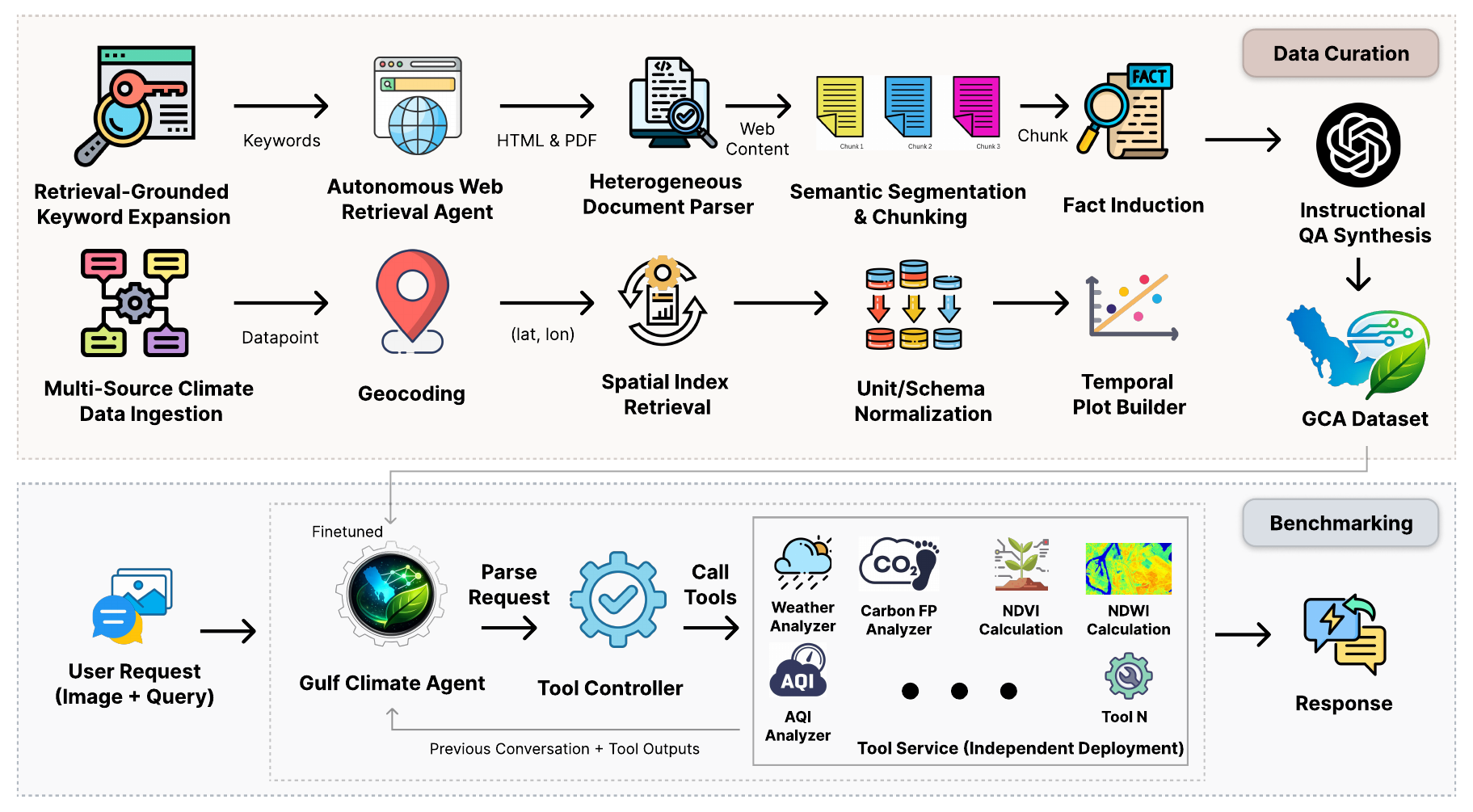}
  \caption{Gulf Climate Agent (GCA) framework. The figure summarizes multimodal dataset curation for text and visual-temporal sources, producing the gca-ds dataset, and the tool-augmented inference stack in which a fine-tuned LLM routes user requests through a tool controller to specialized climate services to generate grounded responses.}
  \label{fig:methodology}
\end{figure*}

\section{GCA Agentic Architecture}
\label{sec:methodology}

Gulf Climate Agent (GCA) is a tool-augmented language agent designed to answer Gulf-focused climate queries by combining (i) domain grounding from our curated multimodal dataset GCA-DS (\S\ref{sec:dataset}) and (ii) structured access to specialised climate analytics tools as shown in Figure \ref{fig:methodology}. Rather than relying on parametric knowledge alone, GCA follows an \emph{act--observe--reason} paradigm in which the model selects tools, consumes their outputs, and iteratively refines intermediate hypotheses until it can produce an evidence-grounded response \citep{yao2023react}.

\subsection{Tool Suite Overview}
\label{subsec:tool-suite}

To keep the interface compact while covering Gulf climate objectives, we group tools into six categories (full signatures in Appendix~A.1): \textbf{Remote Sensing and Land Surface} (satellite retrieval, NDVI/NDWI, desertification change analysis), \textbf{Biodiversity and Species} (bird-call and species recognition), \textbf{Web Retrieval and Summarization} (targeted search and policy/event summarization), \textbf{Carbon and Sustainability} (country/sector footprint estimation), \textbf{Air Quality and Atmospheric Composition} (AQI inquiry/forecasting/trends plus UV and pollen), and \textbf{Weather, Climate, and Hydrology} (weather/rain inquiry and forecasting, river discharge for floods, with geocoding for location resolution). This abstraction supports concise prompting and systematic evaluation of \emph{tool invocation} as a first-class capability.

\subsection{Binding Tools to the LLM}
\label{subsec:binding}

Each tool is exposed to the LLM through a structured function signature (name, arguments, and return schema). At inference time, the model receives: (i) the user query, (ii) brief tool descriptions grouped by category, and (iii) a specification requiring \emph{typed} tool calls. The model produces either a direct answer or a tool call with arguments. Tool outputs are returned verbatim (plus metadata such as units, timestamps, and location), and are appended to the model context as \emph{observations}.

To reduce brittle behavior, we standardize outputs across tools (e.g., normalized units, consistent timestamp formats, and explicit uncertainty when available). This allows the agent to compose multiple tools in a single query (e.g., geocode $\rightarrow$ weather\_analysis $\rightarrow$ summarize) without ad hoc parsing.

\subsection{Agentic Reasoning and Control}
\label{subsec:agentic}

GCA adopts an iterative control loop for tool-augmented reasoning. Given an input query $x$, the agent maintains a trajectory of intermediate steps $\mathcal{T}=\{(a_t,o_t)\}_{t=1}^T$, where $a_t$ is an action (tool call or final response) and $o_t$ is the resulting observation. At each step, the agent first infers the dominant intent, \emph{textual} (policy/event reporting), \emph{numerical} (time-series inquiry), \emph{geospatial} (satellite-derived indices), or \emph{health/environmental} (air quality, UV, pollen), and routes to a minimal set of tool categories. It then selects a tool and arguments, executes the call, and validates schema and units; if the query requires multi-step computation, it chains additional tools and aggregates the resulting evidence. Finally, it synthesizes a response grounded in tool outputs, briefly explaining derived quantities and, when relevant, summarizing temporal trends.

\begin{table}[t]
\centering
\scriptsize
\setlength{\tabcolsep}{2.5pt}
\renewcommand{\arraystretch}{1.05}
\begin{tabularx}{\columnwidth}{lCCC}
\hline
\textbf{Model} & \textbf{Format Err. (\%)} & \textbf{Arg. Err. (\%)} & \textbf{N/A (\%)} \\
\hline
GPT-5             & \cellcolor{red!19}19.8 & \cellcolor{red!12}11.9 & \cellcolor{red!6}6.4 \\
Claude 4.5 Sonnet & \cellcolor{red!19}18.6 & \cellcolor{red!15}14.7 & \cellcolor{red!7}7.1 \\
Gemini 2.5 Pro    & \cellcolor{red!21}21.3 & \cellcolor{red!17}16.5 & \cellcolor{red!7}6.8 \\
Qwen2.5-VL 7B     & \cellcolor{red!47}46.7 & \cellcolor{red!38}38.2 & \cellcolor{red!22}21.5 \\
Pixtral-12B       & \cellcolor{red!39}39.4 & \cellcolor{red!30}29.8 & \cellcolor{red!15}15.2 \\
\hline
\textbf{GCA (ours)} & \cellcolor{red!13}\textbf{12.7} & \cellcolor{red!8}\textbf{8.3} & \cellcolor{red!4}\textbf{3.9} \\
\hline
\end{tabularx}
\vspace{2pt}
\caption{Percentages of tool-use error types on GCA. Format errors: invalid tool-call structure; argument errors: incorrect/missing schema fields; N/A: no actionable tool call when required}
\label{tab:gca-error-types}
\end{table}

\begin{table*}[!htp]
\centering
\small
\setlength{\tabcolsep}{4pt}
\renewcommand{\arraystretch}{1.10}
\begin{tabular}{lcccccc}
\toprule
\textbf{Model} &
\multicolumn{4}{c}{\textbf{Step-by-Step Mode} $\uparrow$} &
\multicolumn{2}{c}{\textbf{End-to-End Mode} $\uparrow$} \\
\cmidrule(lr){2-5}\cmidrule(lr){6-7}
& \textbf{InstAcc} & \textbf{ToolAcc} & \textbf{ArgAcc} & \textbf{SummAcc} & \textbf{AnsAcc} & \textbf{AnsAcc+I} \\
\midrule
GPT-5             & 88.6 & 92.3 & \textbf{90.2} & 87.4 & 86.3 & 87.0 \\
Claude 4.5 Sonnet & 86.8 & 90.2 & 88.3          & 85.6 & 84.3 & 84.8 \\
Gemini 2.5 Pro    & 85.9 & 89.4 & 87.2          & 84.7 & 83.2 & 84.1 \\
Qwen2.5-VL 7B     & 60.5 & 62.2 & 58.3          & 55.6 & 52.3 & 54.0 \\
Pixtral-12B       & 66.9 & 68.3 & 64.2          & 60.8 & 56.2 & 58.5 \\
\midrule
\textbf{GCA (ours)} & \textbf{89.4} & \textbf{94.2} & 89.3 & \textbf{88.6} & \textbf{88.2} & \textbf{89.1} \\
\bottomrule
\end{tabular}
\vspace{2pt}
\caption{Tool-use benchmark results on GCA. Step-by-step mode reports instruction-following accuracy (InstAcc), tool selection accuracy (ToolAcc), argument name accuracy (ArgAcc), and step-conditioned summary accuracy (SummAcc). End-to-end mode reports final execution answer accuracy (AnsAcc) and answer accuracy with image generation enabled (AnsAcc+I).}
\label{tab:gca-tool-benchmark}
\end{table*}

\section{Experiments and Results}
\label{sec:experiments}

We evaluate Gulf Climate Agent (GCA) along two axes: (i) \textbf{domain adaptation} via a single parameter-efficient fine-tuning run on our Gulf-focused multimodal dataset (\S\ref{sec:dataset}), and (ii) \textbf{agentic competence} via a regression benchmark that probes structured tool use under a standardized agent interface. Concretely, we assess whether a model can (a) follow tool-call formatting, (b) choose the correct tool, (c) supply valid arguments, and (d) synthesize a faithful response from tool outputs.

\subsection{Fine-tuning Setup}
\label{subsec:finetune-setup}
We adopt \textbf{Qwen2.5-VL 7B} as the central backbone and apply LoRA-based parameter-efficient fine-tuning \citep{hu2022lora}. We use AdamW with 8-bit optimization, a cosine learning-rate schedule, weight decay $0.1$, learning rate $5\times10^{-5}$ (embedding LR $1\times10^{-5}$), and LoRA settings $r{=}16$, $\alpha{=}16$ targeting \texttt{q\_proj, k\_proj, v\_proj, o\_proj, gate\_proj, up\_proj, down\_proj}. We train on a unified instructional format spanning MCQ, open-ended, and true/false questions across both textual and visual splits, optimizing next-token prediction over formatted samples. This setup exposes the model to Gulf-specific variables (e.g., heat, dust-related air-quality signals, rainfall and discharge) and to tool-facing interaction patterns (location/time/units). As a result, fine-tuning targets both \emph{domain grounding} and \emph{tool-execution reliability} rather than general QA alone.

\subsection{Agentic Benchmark for Tool Usage}
\label{subsec:agent-benchmark}
To test whether models reliably \emph{use} Gulf-relevant climate tools (rather than only describing them), we construct a regression benchmark over the final tool suite (Appendix~\ref{app:appendix}). Each instance provides a user query, tool signatures, and a gold tool-usage trace with one or more calls. The benchmark emphasizes multi-step workflows typical of Gulf climate analysis, such as \texttt{geocode\_mapping} followed by temporal \texttt{weather\_analysis}, or two-date satellite retrieval followed by index computation and change analysis.

\paragraph{Evaluation modes and metrics.}
We report results in two complementary evaluation modes (Table~\ref{tab:gca-tool-benchmark}). \textbf{Step-by-step mode} evaluates each step against gold traces: \textbf{InstAcc} (instruction-following for the step), \textbf{ToolAcc} (correct tool choice), \textbf{ArgAcc} (correct argument \emph{names}/schema fields), and \textbf{SummAcc} (step-conditioned summary correctness given tool outputs). \textbf{End-to-end mode} evaluates the full execution outcome: \textbf{AnsAcc} measures final answer accuracy after executing the predicted tool trace, and \textbf{AnsAcc+I} enables image generation during response composition (useful when answers require visual explanation of temporal trends).

\paragraph{Error taxonomy.}
To diagnose failures, Table~\ref{tab:gca-error-types} summarizes tool-use error rates: \textbf{Format Err.} (invalid tool-call structure), \textbf{Arg. Err.} (incorrect or missing schema fields), and \textbf{N/A} (no actionable tool call when a tool is required). These error types map directly to common agent breakdowns in practice: format errors prevent execution entirely, argument errors yield invalid or misleading tool runs, and N/A reflects premature ``direct answering'' without tool grounding.


\noindent\textbf{Baselines.} We compare against representative proprietary and open models under identical prompting, tool schemas, and execution harness: \textbf{GPT-5}, \textbf{Claude 4.5 Sonnet}, \textbf{Gemini 2.5 Pro}, \textbf{Qwen2.5-VL 7B}, and \textbf{Pixtral-12B} (\texttt{mistralai/pixtral-12b}). All models operate in a tool-augmented setting where they may either answer directly or emit structured tool calls.

\subsection{Main Results}
\label{subsec:main-results}
Table~\ref{tab:gca-tool-benchmark} reports the main benchmark results. Overall, \textbf{GCA is best or competitive across all metrics} and substantially improves tool reliability relative to its base backbone. In \textbf{step-by-step mode}, GCA attains the highest \textbf{ToolAcc} (94.2) and \textbf{SummAcc} (88.6), with strong \textbf{InstAcc} (89.4) and \textbf{ArgAcc} (89.3). In \textbf{end-to-end mode}, GCA achieves the highest \textbf{AnsAcc} (88.2) and \textbf{AnsAcc+I} (89.1), indicating that the gains in structured execution translate into improved final answers rather than only cleaner traces.

\paragraph{Effect of fine-tuning.}
Comparing GCA to the \emph{unadapted} Qwen2.5-VL 7B baseline highlights the impact of a single LoRA run on tool competence. GCA improves by \textbf{+32.0} ToolAcc (94.2 vs.\ 62.2), \textbf{+31.0} ArgAcc (89.3 vs.\ 58.3), and \textbf{+33.0} SummAcc (88.6 vs.\ 55.6). These step-level gains compound into a large end-to-end improvement of \textbf{+35.9} AnsAcc (88.2 vs.\ 52.3) and \textbf{+35.1} AnsAcc+I (89.1 vs.\ 54.0). The error breakdown in Table~\ref{tab:gca-error-types} is consistent with this effect: relative to Qwen2.5-VL 7B, GCA sharply reduces \textbf{Format Err.} (12.7\% vs.\ 46.7\%), \textbf{Arg. Err.} (8.3\% vs.\ 38.2\%), and \textbf{N/A} (3.9\% vs.\ 21.5\%). This indicates that fine-tuning primarily improves (i) emitting executable tool calls and (ii) adhering to schema constraints, which are prerequisites for successful multi-step tool composition.

\paragraph{Comparison to strong proprietary models.}
GCA remains competitive with proprietary baselines. Against GPT-5, GCA improves \textbf{ToolAcc} by \textbf{+1.9} (94.2 vs.\ 92.3), \textbf{SummAcc} by \textbf{+1.2} (88.6 vs.\ 87.4), and \textbf{AnsAcc} by \textbf{+1.9} (88.2 vs.\ 86.3), while staying close on argument naming accuracy (89.3 vs.\ 90.2). Claude 4.5 Sonnet and Gemini 2.5 Pro trail GCA more clearly on end-to-end accuracy (84.3/83.2 vs.\ 88.2). Importantly, Table~\ref{tab:gca-error-types} shows that GCA also exhibits the lowest error rates among all evaluated models, suggesting that Gulf-specific training improves not only answer quality but also execution robustness.

\paragraph{Why do scores improve?}
We provide two quantitative signals that the gains are driven by improved \emph{execution behavior} rather than stylistic differences. 
First, Table~\ref{tab:gca-error-types} shows that GCA substantially reduces tool-call failures: format errors drop to 12.7\% (vs.\ 46.7\% for Qwen2.5-VL 7B; 19.8--21.3\% for GPT-5/Gemini), argument errors drop to 8.3\% (vs.\ 38.2\%), and ``no-action'' cases drop to 3.9\% (vs.\ 21.5\%). These reductions explain higher end-to-end accuracy because invalid or missing calls cannot be executed and therefore cannot yield grounded answers.
Second, Table~\ref{tab:gca-tool-benchmark} indicates that improvements concentrate on the steps most sensitive to tool execution: relative to Qwen2.5-VL 7B, GCA improves \textbf{ToolAcc} (+32.0) and \textbf{ArgAcc} (+31.0), which in turn raises \textbf{SummAcc} (+33.0) and final \textbf{AnsAcc} (+35.9). Together, the lower error rates and the step-level gains provide direct empirical evidence that fine-tuning improves structured tool use (formatting + schema adherence), which is the dominant driver of downstream answer improvements.

\section{Conclusion}
\label{sec:conclusion}

We introduced \textbf{Gulf Climate Agent (GCA)}, a Gulf-focused, tool-augmented climate assistant that bridges general-purpose LLM/VLM reasoning with regional climate evidence and specialised analytics. The proposed \textbf{GCA framework} contributes (i) a semi-automated curation pipeline producing \textbf{GCA-DS}, a \textbf{multimodal} Gulf dataset of \textbf{$\sim$200k} QA pairs spanning policies, reports, literature, event news, and city-level visual-temporal variables; (ii) a modular tool suite for remote sensing, air quality/health indices, weather/rainfall, hydrology, carbon, web retrieval/summarization, and geocoding; and (iii) a benchmarking and fine-tuning study showing improved tool-use reliability and end-to-end answer accuracy over strong baselines. Future work will expand continual data ingestion, strengthen validation, and add uncertainty-aware and scenario-driven analyses for Gulf climate objectives.

\clearpage

\section{Limitations}
\label{sec:limitations}

GCA is designed for Gulf-focused climate decision support, but it has several limitations. 
First, \textbf{coverage and validation} of GCA-DS are semi validated: although the dataset is large and semi-automatically curated, only a subset is human-verified, and automated QA generation can introduce subtle label or grounding errors (especially for long policy documents and event reports). 
Second, \textbf{tool dependence} constrains reliability: the agent’s outputs inherit biases, resolution limits, and missing-data patterns from upstream data products and analytical modules (e.g., spatiotemporal gaps, differing units/definitions across sources). Tool failures can also propagate through multi-step traces, even when the language model’s intent is correct. 
Third, \textbf{benchmark scope} is necessarily selective: our regression suite emphasizes Gulf-relevant tool workflows and does not exhaustively cover all climate tasks (e.g., full uncertainty quantification, attribution, or long-horizon scenario planning). 
Finally, \textbf{deployment considerations} remain: real-world use requires careful monitoring, transparent provenance, and domain-expert oversight, particularly for high-stakes recommendations where uncertainty and stakeholder constraints must be explicitly communicated.

\clearpage


\bibliography{custom}


\newcommand{\toolname}[2]{\begin{tabular}[t]{@{}l@{}}\texttt{#1}\\\texttt{#2}\end{tabular}}

\appendix
\section{Appendix}
\subsection{Tool Suite}
\label{app:appendix}


\begin{table*}[!t]
\centering
\footnotesize
\setlength{\tabcolsep}{5pt}
\renewcommand{\arraystretch}{1.2}
\begin{tabular}{p{2.7cm} p{6.3cm} p{2.8cm} p{2.8cm}}
\toprule
\textbf{Tool} & \textbf{Purpose} & \textbf{Inputs} & \textbf{Outputs} \\
\midrule

\multicolumn{4}{l}{\textbf{Remote sensing and land surface}} \\
\toolname{get\_satellite\_}{image} &
Retrieve a multispectral satellite image for a coordinate and date, used for downstream index and change analyses. &
\texttt{lat, lon, date} &
\texttt{image} + metadata. \\

\toolname{calculate\_}{ndvi} &
Compute NDVI from an image to quantify vegetation condition (scalar map + summary). &
\texttt{image} &
\texttt{ndvi\_map} + stats. \\

\toolname{calculate\_}{ndwi} &
Compute NDWI from an image to highlight water/moisture signals (scalar map + summary). &
\texttt{image} &
\texttt{ndwi\_map} + stats. \\

\toolname{desertification\_}{analysis} &
Compare two images and return land-degradation indicators (e.g., index deltas and affected area). &
\texttt{image1, image2} &
\texttt{change\_map} + metrics. \\

\midrule
\multicolumn{4}{l}{\textbf{Biodiversity and species}} \\
\toolname{detect\_}{bird} &
Recognize bird calls from audio, returning candidate species with confidence. &
\texttt{audio\_clip} &
(\texttt{species, conf}) list. \\

\toolname{detect\_}{species} &
Classify plant/animal species from an image, returning candidates with confidence. &
\texttt{image} &
(\texttt{species, conf}) list. \\

\midrule
\multicolumn{4}{l}{\textbf{Web retrieval and summarization}} \\
\toolname{online\_}{search} &
Targeted search for policies, reports, and event coverage; returns ranked results with snippets. &
\texttt{query} &
(\texttt{title,url,snippet}) list. \\

\toolname{summa}{rize} &
Produce a concise summary preserving key facts and implications. &
\texttt{text} &
\texttt{summary}. \\

\midrule
\multicolumn{4}{l}{\textbf{Carbon and sustainability}} \\
\toolname{carbon\_footprint\_}{calculation} &
Estimate annual emissions for a country/industry/year given revenue (factor-based). &
\texttt{country, industry, year, revenue} &
\texttt{tCO$_2$e}. \\

\midrule
\multicolumn{4}{l}{\textbf{Air quality and health indices}} \\
\toolname{aqi\_}{inquiry} &
Return AQI and pollutant values for a location and date. &
\texttt{lat, lon, date} &
\texttt{aqi} + pollutants. \\

\toolname{aqi\_}{prediction} &
Forecast AQI for a location over a specified horizon. &
\texttt{lat, lon, horizon} &
AQI time series. \\

\toolname{aqi\_}{analysis} &
Summarize AQI trends and exceedances over a date range. &
\texttt{lat, lon, start, end} &
Stats + trend + exceedances. \\

\toolname{pollen\_}{forecast} &
Return forecast pollen levels for a location. &
\texttt{lat, lon} &
\texttt{pollen\_levels}. \\

\toolname{uv\_index\_}{forecast} &
Return UV index forecast for a location. &
\texttt{lat, lon} &
UV time series. \\

\midrule
\multicolumn{4}{l}{\textbf{Weather, rainfall, and hydrology}} \\
\toolname{weather\_}{inquiry} &
Return historical weather variables for a location and date. &
\texttt{lat, lon, date} &
Weather dict. \\

\toolname{weather\_}{forecast} &
Return weather forecast for the next $n$ days. &
\texttt{lat, lon, days} &
Forecast series. \\

\toolname{weather\_}{analysis} &
Compute summary statistics and anomalies over a date range. &
\texttt{lat, lon, start, end} &
Stats + anomalies. \\

\toolname{rain\_}{inquiry} &
Return precipitation for a location and date. &
\texttt{lat, lon, date} &
\texttt{precip} (mm). \\

\toolname{rain\_}{prediction} &
Forecast precipitation for a location over a specified horizon. &
\texttt{lat, lon, horizon} &
Precip series. \\

\toolname{rain\_}{analysis} &
Summarize rainfall patterns and extremes over a date range. &
\texttt{lat, lon, start, end} &
Stats + events. \\

\toolname{river\_discharge\_}{check} &
Return simulated river discharge for the nearest river grid cell at a date. &
\texttt{lat, lon, date} &
\texttt{discharge} (m$^3$/s). \\

\midrule
\multicolumn{4}{l}{\textbf{Geospatial utility}} \\
\toolname{geocode\_}{mapping} &
Resolve a region/city name to coordinates for downstream tool calls. &
\texttt{region} &
\texttt{lat, lon} (+ metadata). \\

\bottomrule
\end{tabular}
\vspace{2pt}
\caption{\textbf{Tool Suite Details.} GCA tool suite. Each tool is exposed to the LLM with typed inputs/outputs; outputs are normalized (units, timestamps) to support multi-step composition.}
\label{tab:tools}
\end{table*}

\end{document}